\documentclass[10pt,twocolumn,letterpaper]{article}

\usepackage{cvpr}
\usepackage{times}
\usepackage{epsfig}
\usepackage{graphicx}
\usepackage{amsmath}
\usepackage{amssymb}
\usepackage{subcaption}

\usepackage{multirow}
\usepackage{color}

\usepackage[pagebackref=true,breaklinks=true,letterpaper=true,colorlinks,bookmarks=false]{hyperref}

\cvprfinalcopy 

\def\cvprPaperID{1260} 

\ifcvprfinal\fi
\begin{document}

\title{Feature Selective Anchor-Free Module for Single-Shot Object Detection}

\author{Chenchen Zhu \qquad Yihui He \qquad Marios Savvides\\
Carnegie Mellon University\\
{\tt\small \{chenchez, he2, marioss\}@andrew.cmu.edu}
}

\twocolumn[{%
\maketitle
\ifcvprfinal\thispagestyle{empty}\fi
\begin{center}
        \includegraphics[width=0.47\textwidth]{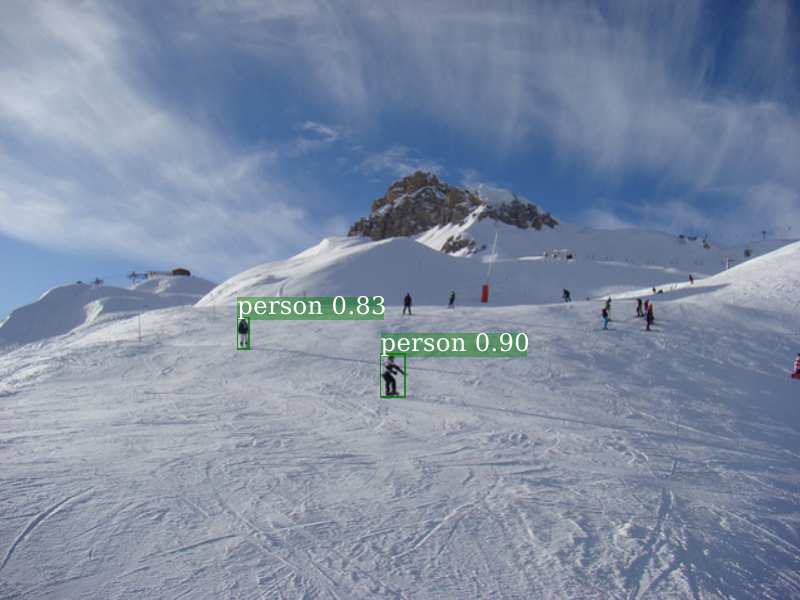}
        \includegraphics[width=0.47\textwidth]{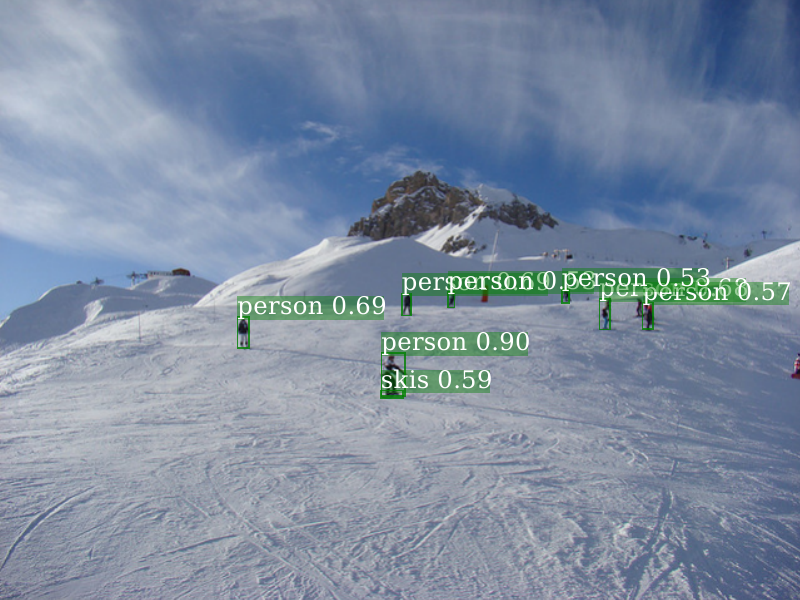}
        \captionof{subfigure}{RetinaNet (anchor-based, ResNeXt-101) \qquad \qquad \qquad b: Ours (anchor-based + FSAF, ResNet-50)}
    \captionof{figure}{Qualitative results of the anchor-based RetinaNet~\cite{retinanet} using powerful \textit{ResNeXt-101} (left) and our detector with additional FSAF module using just \textit{ResNet-50} (right) under the same training and testing scale. Our FSAF module helps detecting hard objects like tiny person and flat skis with a less powerful backbone network. See Figure~\ref{fig:ablation:compare} for more examples.}
    \label{fig:compare}
\end{center}%
}]

\begin{abstract}
   We motivate and present feature selective anchor-free (FSAF) module, a simple and effective building block for single-shot object detectors. It can be plugged into single-shot detectors with feature pyramid structure. The FSAF module addresses two limitations brought up by the conventional anchor-based detection: 1) heuristic-guided feature selection; 2) overlap-based anchor sampling. The general concept of the FSAF module is online feature selection applied to the training of multi-level anchor-free branches. Specifically, an anchor-free branch is attached to each level of the feature pyramid, allowing box encoding and decoding in the anchor-free manner at an arbitrary level. During training, we dynamically assign each instance to the most suitable feature level. At the time of inference, the FSAF module can work jointly with anchor-based branches by outputting predictions in parallel. We instantiate this concept with simple implementations of anchor-free branches and online feature selection strategy. Experimental results on the COCO detection track show that our FSAF module performs better than anchor-based counterparts while being faster. When working jointly with anchor-based branches, the FSAF module robustly improves the baseline RetinaNet by a large margin under various settings, while introducing nearly free inference overhead. And the resulting best model can achieve a state-of-the-art 44.6\% mAP, outperforming all existing single-shot detectors on COCO.
\end{abstract}

\section{Introduction}

\begin{figure}
    \centering
    \includegraphics[width=\columnwidth]{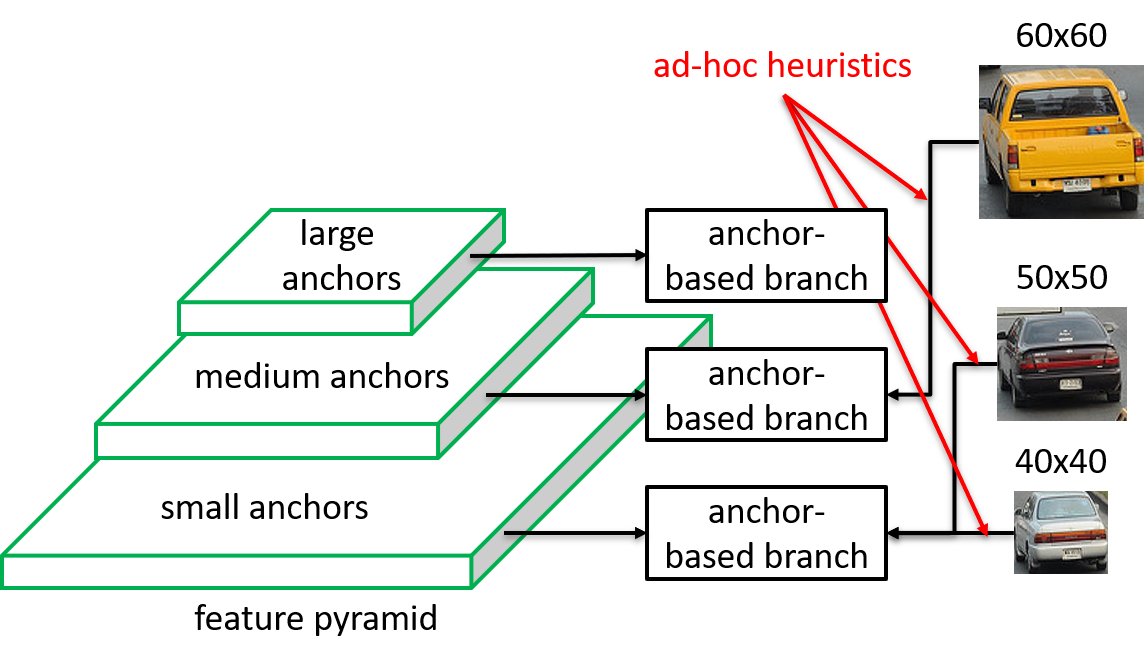}
    \caption{Selected feature level in anchor-based branches may not be optimal.}
    \label{fig:intro}
\end{figure}

\begin{figure*}
    \centering
    \includegraphics[width=1.8\columnwidth]{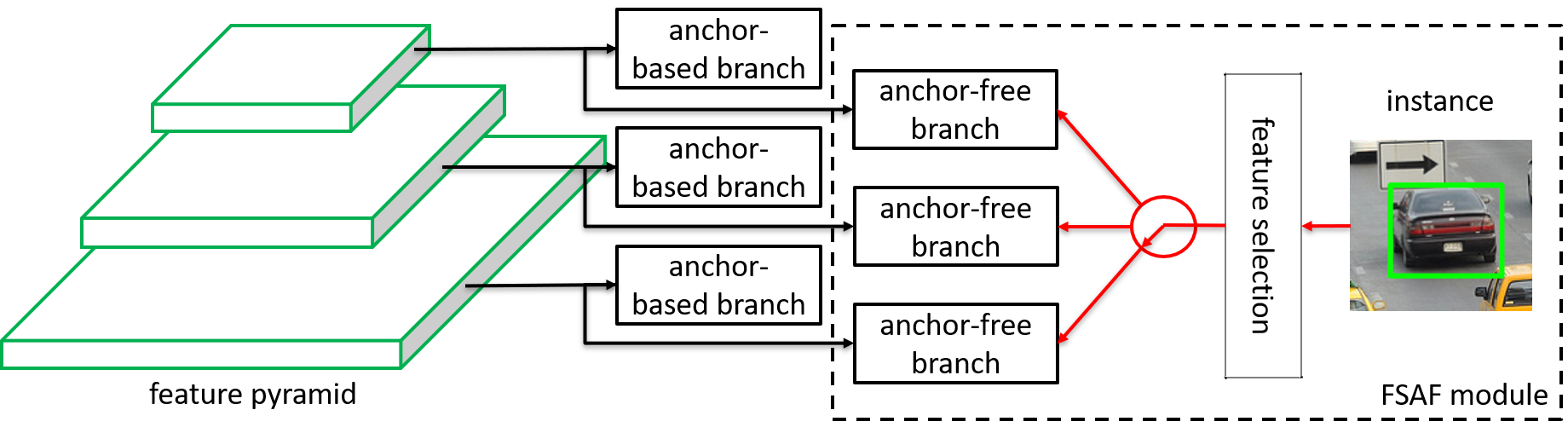}
    \caption{Overview of our FSAF module plugged into conventional anchor-based detection methods. During training, each instance is assigned to a pyramid level via feature selection for setting up supervision signals.}
    \label{fig:architecture}
\end{figure*}

Object detection is an important task in the computer vision community. It serves as a prerequisite for various downstream vision applications such as instance segmentation~\cite{mask}, facial analysis~\cite{3dtps, ringloss}, autonomous driving cars~\cite{dollar2009pedestrian, liang2018cirl}, and video analysis~\cite{ma2018vehicle, videographs}. The performance of object detectors has been dramatically improved thanks to the advance of deep convolutional neural networks~\cite{alexnet, vgg, resnet, resnext} and well-annotated datasets~\cite{voc, coco}. 

One challenging problem for object detection is scale variation. To achieve scale invariability, state-of-the-art detectors construct feature pyramids or multi-level feature towers~\cite{ssd, dssd, fpn, retinanet, detnet, m2det}. And multiple scale levels of feature maps are generating predictions in parallel. Besides, anchor boxes can further handle scale variation~\cite{ssd, fasterrcnn}. Anchor boxes are designed for discretizing the continuous space of all possible instance boxes into a finite number of boxes with predefined locations, scales and aspect ratios. And instance boxes are matched to anchor boxes based on the Intersection-over-Union (IoU) overlap. When integrated with feature pyramids, large anchor boxes are typically associated with upper feature maps, and small anchor boxes are associated with lower feature maps, see Figure~\ref{fig:intro}. This is based on the heuristic that upper feature maps have more semantic information suitable for detecting big instances whereas lower feature maps have more fine-grained details suitable for detecting small instances~\cite{hypercolumn}. The design of feature pyramids integrated with anchor boxes has achieved good performance on object detection benchmarks~\cite{voc, coco, kitti}.

However, this design has two limitations: 1) heuristic-guided feature selection; 2) overlap-based anchor sampling. During training, each instance is always matched to the closest anchor box(es) according to IoU overlap. And anchor boxes are associated with a certain level of feature map by human-defined rules, such as box size. Therefore, the selected feature level for each instance is purely based on \textit{ad-hoc heuristics}. For example, a car instance with size $50\times50$ pixels and another similar car instance with size $60\times60$ pixels may be assigned to two different feature levels, whereas another $40\times40$ car instance may be assigned to the same level as the $50\times50$ instance, as illustrated in Figure~\ref{fig:intro}. In other words, the anchor matching mechanism is inherently heuristic-guided. This leads to a major flaw that the selected feature level to train each instance may not be optimal.

We propose a simple and effective approach named feature selective anchor-free (FSAF) module to address these two limitations simultaneously. Our motivation is to let each instance select the best level of feature freely to optimize the network, so there should be no anchor boxes to constrain the feature selection in our module. Instead, we encode the instances in an anchor-free manner to learn the parameters for classification and regression. The general concept is presented in Figure~\ref{fig:architecture}. An anchor-free branch is built per level of feature pyramid, independent to the anchor-based branch. Similar to the anchor-based branch, it consists of a classification subnet and a regression subnet (not shown in figure). An instance can be assigned to arbitrary level of the anchor-free branch. During training, we dynamically select the most suitable level of feature for each instance based on the instance content instead of just the size of instance box. The selected level of feature then learns to detect the assigned instances. At inference, the FSAF module can run independently or jointly with anchor-based branches. Our FSAF module is agnostic to the backbone network and can be applied to single-shot detectors with a structure of feature pyramid. Additionally, the instantiation of anchor-free branches and online feature selection can be various. In this work, we keep the implementation of our FSAF module simple so that its computational cost is marginal compared to the whole network.

Extensive experiments on the COCO~\cite{coco} object detection benchmark confirm the effectiveness of our method. The FSAF module by itself outperforms anchor-based counterparts as well as runs faster. When working jointly with anchor-based branches, the FSAF module can consistently improve the strong baselines by large margins across various backbone networks, while at the same time introducing the minimum cost of computation. Especially, we improve RetinaNet using ResNeXt-101~\cite{resnext} by \textbf{1.8\%} with only \textbf{6ms} additional inference latency. 
Additionally, our final detector achieves a state-of-the-art \textbf{44.6}\% mAP when multi-scale testing are employed, outperforming all existing single-shot detectors on COCO.


\section{Related Work}
Recent object detectors often use feature pyramid or multi-level feature tower as a common structure. SSD~\cite{ssd} first proposed to predict class scores and bounding boxes from multiple feature scales. FPN~\cite{fpn} and DSSD~\cite{dssd} proposed to enhance low-level features with high-level semantic feature maps at all scales. RetinaNet~\cite{retinanet} addressed class imbalance issue of multi-level dense detectors with focal loss. DetNet~\cite{detnet} designed a novel backbone network to maintain high spatial resolution in upper pyramid levels. However, they all use pre-defined anchor boxes to encode and decode object instances. Other works address the scale variation differently. Zhu et al~\cite{zccfd} enhanced the anchor design for small objects. He et al~\cite{softernms} modeled the bounding box as Gaussian distribution for improved localization.

The idea of anchor-free detection is not new. DenseBox~\cite{densebox} first proposed a unified end-to-end fully convolutional framework that directly predicted bounding boxes. UnitBox~\cite{unitbox} proposed an Intersection over Union (IoU) loss function for better box regression. Zhong et al~\cite{af-rpn} proposed anchor-free region proposal network to find text in various scales, aspect ratios, and orientations. Recently CornerNet~\cite{cornernet} proposed to detect an object bounding box as a pair of corners, leading to the best single-shot detector. SFace~\cite{sface} proposed to integrate the anchor-based method and anchor-free method. However, they still adopt heuristic feature selection strategies. 


\section{Feature Selective Anchor-Free Module}
In this section we instantiate our feature selective anchor-free (FSAF) module by showing how to apply it to the single-shot detectors with feature pyramids, such as SSD~\cite{ssd}, DSSD~\cite{dssd} and RetinaNet~\cite{retinanet}. Without lose of generality, we apply the FSAF module to the state-of-the-art RetinaNet~\cite{retinanet} and demonstrate our design from the following aspects: 1) how to create the anchor-free branches in the network (\ref{section:method:architecture}); 2) how to generate supervision signals for anchor-free branches (\ref{section:method:label}); 3) how to dynamically select feature level for each instance (\ref{section:method:feature_selection}); 4) how to jointly train and test anchor-free and anchor-based branches (\ref{section:method:training}).

\subsection{Network Architecture}
\label{section:method:architecture}

\begin{figure*}
    \centering
    \includegraphics[width=1.8\columnwidth]{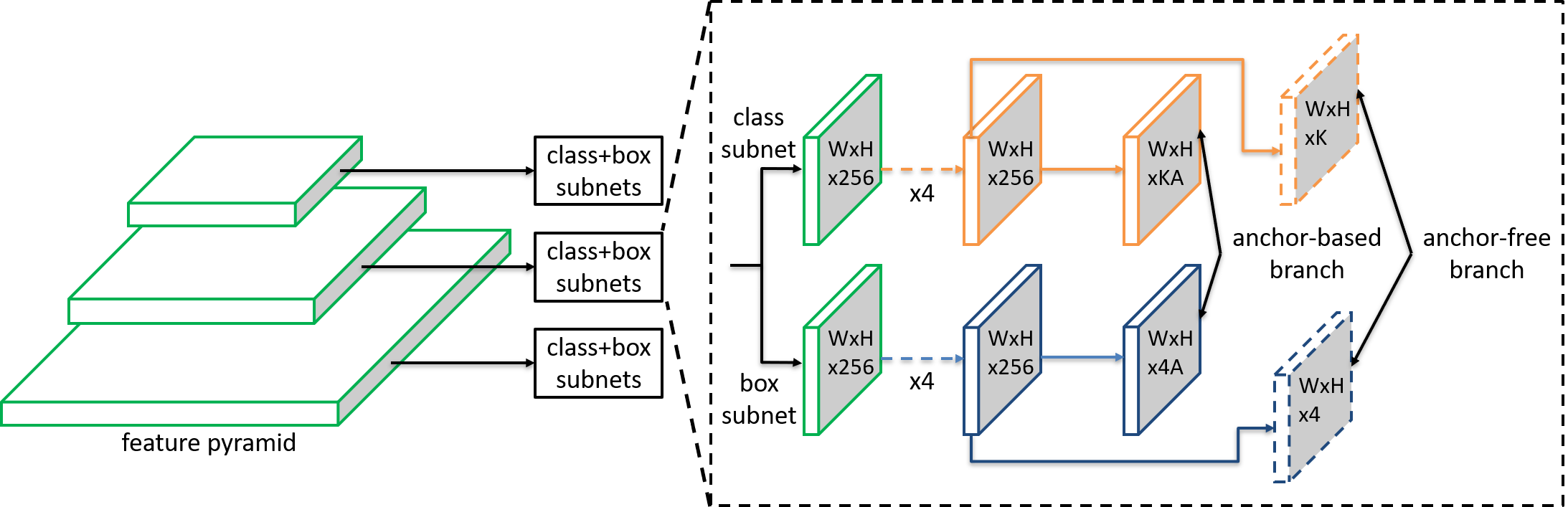}
    \caption{Network architecture of RetinaNet with our FSAF module. The FSAF module only introduces two additional conv layers (dashed feature maps) per pyramid level, keeping the architecture fully convolutional.}
    \label{fig:af_retina}
\end{figure*}

From the network's perspective, our FSAF module is surprisingly simple. Figure~\ref{fig:af_retina} illustrates the architecture of the RetinaNet~\cite{retinanet} with the FSAF module. In brief, RetinaNet is composed of a backbone network (not shown in the figure) and two task-specific subnets. The feature pyramid is constructed from the backbone network with levels from $P_3$ through $P_7$, where $l$ is the pyramid level and $P_l$ has $1/2^l$ resolution of the input image. Only three levels are shown for simplicity. Each level of the pyramid is used for detecting objects at a different scale. To do this, a classification subnet and a regression subnet are attached to $P_l$. They are both small fully convolutional networks. The classification subnet predicts the probability of objects at each spatial location for each of the $A$ anchors and $K$ object classes. The regression subnet predicts the 4-dimensional class-agnostic offset from each of the $A$ anchors to a nearby instance if exists.

On top of the RetinaNet, our FSAF module introduces only two additional conv layers per pyramid level, shown as the dashed feature maps in Figure~\ref{fig:af_retina}. These two layers are responsible for the classification and regression predictions in the anchor-free branch respectively. To be more specific, a $3\times3$ conv layer with $K$ filters is attached to the feature map in the classification subnet followed by the sigmoid function, in parallel with the one from the anchor-based branch. It predicts the probability of objects at each spatial location for $K$ object classes. Similarly, a $3\times3$ conv layer with four filters is attached to the feature map in the regression subnet followed by the ReLU~\cite{relu} function. It is responsible for predicting the box offsets encoded in an anchor-free manner. To this end the anchor-free and anchor-based branches work jointly in a multi-task style, sharing the features in every pyramid level. 

\subsection{Ground-truth and Loss}
\label{section:method:label}

\begin{figure}
    \centering
    \includegraphics[width=0.8\columnwidth]{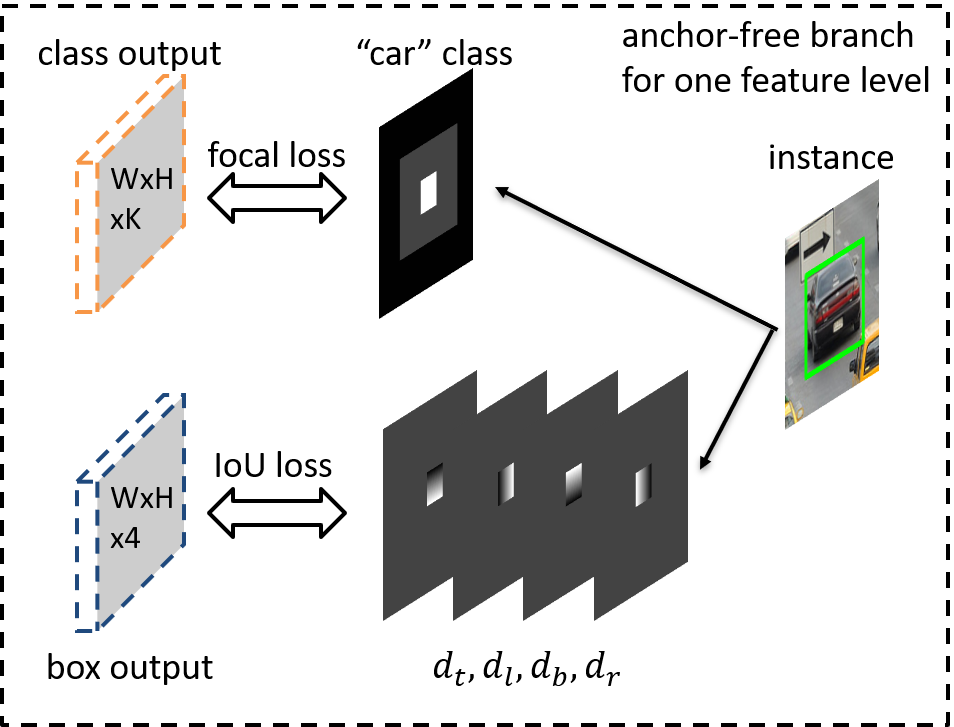}
    \caption{Supervision signals for an instance in one feature level of the anchor-free branches. We use focal loss for classification and IoU loss for box regression.}
    \label{fig:loss}
\end{figure}

Given an object instance, we know its class label $k$ and bounding box coordinates $b = [x, y, w, h]$, where $(x, y)$ is the center of the box, and $w, h$ are box width and height respectively. The instance can be assigned to arbitrary feature level $P_l$ during training. We define the projected box $b_p^l = [x_p^l, y_p^l, w_p^l, h_p^l]$ as the projection of $b$ onto the feature pyramid $P_l$, i.e. $b_p^l = b / 2^l$. We also define the effective box $b_e^l = [x_e^l, y_e^l, w_e^l, h_e^l]$ and the ignoring box $b_i^l = [x_i^l, y_i^l, w_i^l, h_i^l]$ as proportional regions of $b_p^l$ controlled by constant scale factors $\epsilon_e$ and $\epsilon_i$ respectively, i.e. $x_e^l = x_p^l, y_e^l = y_p^l, w_e^l = \epsilon_e w_p^l, h_e^l = \epsilon_e h_p^l$, $x_i^l = x_p^l, y_i^l = y_p^l, w_i^l = \epsilon_i w_p^l, h_i^l = \epsilon_i h_p^l$. We set $\epsilon_e = 0.2$ and $\epsilon_i = 0.5$. An example of ground-truth generation for a car instance is illustrated in Figure~\ref{fig:loss}.

\noindent \textbf{Classification Output:} The ground-truth for the classification output is $K$ maps, with each map corresponding to one class. The instance affects $k$th ground-truth map in three ways. First, the effective box $b_e^l$ region is the positive region filled by ones shown as the white box in ``car'' class map, indicating the existence of the instance. Second, the ignoring box excluding the effective box ($b_i^l - b_e^l$) is the ignoring region shown as the grey area, which means that the gradients in this area are not propagated back to the network. Third, the ignoring boxes in adjacent feature levels ($b_i^{l-1}$, $b_i^{l+1}$) are also ignoring regions if exists. Note that if the effective boxes of two instances overlap in one level, the smaller instance has higher priority.
The rest region of the ground-truth map is the negative (black) area filled by zeros, indicating the absence of objects. Focal loss~\cite{retinanet} is applied for supervision with hyperparameters $\alpha = 0.25$ and $\gamma = 2.0$. The total classification loss of anchor-free branches for an image is the summation of the focal loss over all non-ignoring regions, normalized by the total number of pixels inside all effective box regions.

\noindent \textbf{Box Regression Output:} The ground-truth for the regression output are 4 offset maps agnostic to classes. The instance only affects the $b_e^l$ region on the offset maps. For each pixel location $(i, j)$ inside $b_e^l$, we represent the projected box $b_p^l$ as a 4-dimensional vector $\mathbf{d}_{i,j}^l = [d_{t_{i,j}}^l, d_{l_{i,j}}^l, d_{b_{i,j}}^l, d_{r_{i,j}}^l]$, where $d_t^l$, $d_l^l$, $d_b^l$, $d_r^l$ are the distances between the current pixel location $(i, j)$ and the top, left, bottom, and right boundaries of $b_p^l$, respectively. Then the 4-dimensional vector at $(i, j)$ location across 4 offset maps is set to $\mathbf{d}_{i,j}^l / S$ with each map corresponding to one dimension. $S$ is a normalization constant and we choose $S=4.0$ in this work empirically. Locations outside the effective box are the grey area where gradients are ignored. IoU loss~\cite{unitbox} is adopted for optimization. The total regression loss of anchor-free branches for an image is the average of the IoU loss over all effective box regions.

During inference, it is straightforward to decode the predicted boxes from the classification and regression outputs. At each pixel location $(i, j)$, suppose the predicted offsets are $[\hat{o}_{t_{i,j}}, \hat{o}_{l_{i,j}}, \hat{o}_{b_{i,j}}, \hat{o}_{r_{i,j}}]$. Then the predicted distances are $[S\hat{o}_{t_{i,j}}, S\hat{o}_{l_{i,j}}, S\hat{o}_{b_{i,j}}, S\hat{o}_{r_{i,j}}]$. And the top-left corner and the bottom-right corner of the predicted projected box are $(i - S\hat{o}_{t_{i,j}}, j - S\hat{o}_{l_{i,j}})$ and $(i + S\hat{o}_{b_{i,j}}, j + S\hat{o}_{r_{i,j}}])$ respectively. We further scale up the projected box by $2^l$ to get the final box in the image plane. The confidence score and class for the box can be decided by the maximum score and the corresponding class of the K-dimensional vector at location $(i, j)$ on the classification output maps.

\begin{figure*}
    \centering
    \includegraphics[width=1.8\columnwidth]{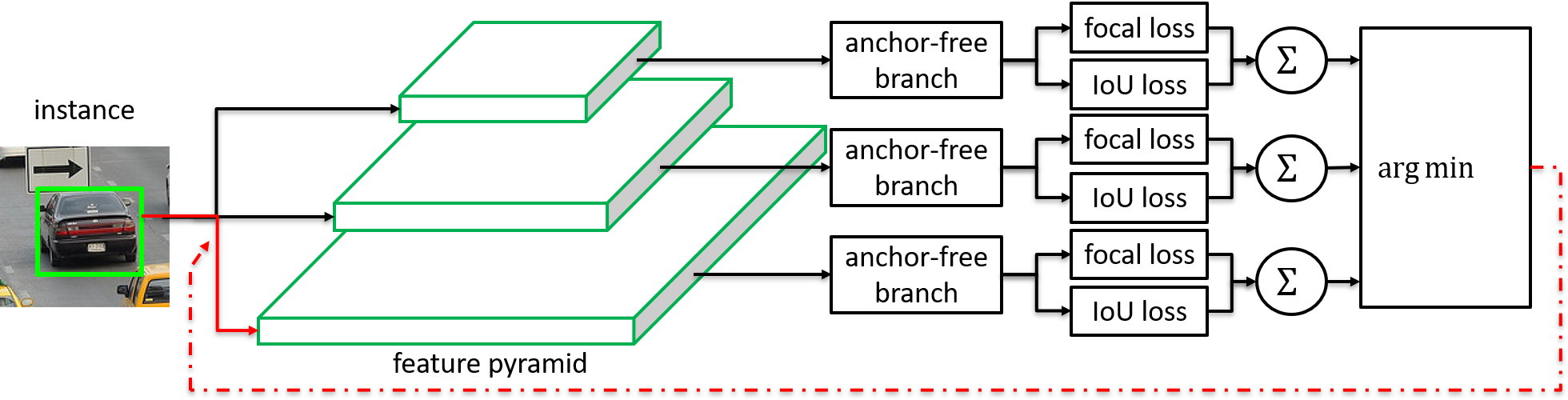}
    \caption{Online feature selection mechanism. Each instance is passing through all levels of anchor-free branches to compute the averaged classification (focal) loss and regression (IoU) loss over effective regions. Then the level with minimal summation of two losses is selected to set up the supervision signals for that instance.}
    \label{fig:feat_select}
\end{figure*}

\subsection{Online Feature Selection}
\label{section:method:feature_selection}

The design of the anchor-free branches allows us to learn each instance using the feature of an arbitrary pyramid level $P_l$. To find the optimal feature level, our FSAF module selects the best $P_l$ based on the instance content, instead of the size of instance box as in anchor-based methods. 

Given an instance $I$, we define its classification loss and box regression loss on $P_l$ as $L_{FL}^I(l)$ and $L_{IoU}^I(l)$, respectively. They are computed by averaging the focal loss and the IoU loss over the effective box region $b_e^l$, i.e.
\begin{align} \label{eq:inst_loss}
\begin{split}
    L_{FL}^I(l) &= \frac{1}{N(b_e^l)} \sum_{i,j \in b_e^l} FL(l, i, j) \\
    L_{IoU}^I(l) &= \frac{1}{N(b_e^l)} \sum_{i,j \in b_e^l} IoU(l, i, j)
\end{split}
\end{align}
where $N(b_e^l)$ is the number of pixels inside $b_e^l$ region, and $FL(l, i, j)$, $IoU(l, i, j)$ are the focal loss~\cite{retinanet} and IoU loss~\cite{unitbox} at location $(i, j)$ on $P_l$ respectively.

Figure~\ref{fig:feat_select} shows our online feature selection process. First the instance $I$ is forwarded through all levels of feature pyramid. Then the summation of $L_{FL}^I(l)$ and $L_{IoU}^I(l)$ is computed in all anchor-free branches using Eqn.~\eqref{eq:inst_loss}. Finally, the best pyramid level $P_{l^*}$ yielding the minimal summation of losses is selected to learn the instance, i.e.
\begin{equation} \label{eq:online_fs}
    l^* = \arg\min_l L_{FL}^I(l) + L_{IoU}^I(l)
\end{equation}
For a training batch, features are updated for their correspondingly assigned instances. The intuition is that the selected feature is currently the best to model the instance. Its loss forms a lower bound in the feature space. And by training, we further pull down this lower bound. At the time of inference, we do not need to select the feature because the most suitable level of feature pyramid will naturally output high confidence scores.

In order to verify the importance of our online feature selection, we also conduct a heuristic feature selection process for comparison in the ablation studies (\ref{section:exp:ablation}). The heuristic feature selection depends purely on box sizes. We borrow the idea from the FPN detector~\cite{fpn}. An instance $I$ is assigned to the level $P_{l'}$ of the feature pyramid by:
\begin{equation} \label{eq:naive_fs}
    l' = \lfloor l_0 + \log_2 (\sqrt{w h} / 224) \rfloor
\end{equation}
Here 224 is the canonical ImageNet pre-training size, and $l_0$ is the target level on which an instance with $w \times h = 224^2$ should be mapped into. In this work we choose $l_0 = 5$ because ResNet~\cite{resnet} uses the feature map from 5th convolution group to do the final classification.

\begin{table*}
\centering
\begin{tabular}{c|c|c c|c c c c c c}
\hline\hline
\multirow{2}{*}{} & \multirow{2}{*}{\begin{tabular}[c]{@{}c@{}}Anchor-\\based\\branches\end{tabular}} & \multicolumn{2}{c|}{Anchor-free branches} & \multirow{3}{*}{AP} & \multirow{3}{*}{AP$_{50}$} & \multirow{3}{*}{AP$_{75}$} & \multirow{3}{*}{AP$_{S}$} & \multirow{3}{*}{AP$_{M}$} & \multirow{3}{*}{AP$_{L}$} \\ \cline{3-4}
 &  & \begin{tabular}[c]{@{}c@{}}Heuristic feature \\ selection Eqn.~\eqref{eq:naive_fs}\end{tabular} & \begin{tabular}[c]{@{}c@{}}Online feature \\ selection Eqn.~\eqref{eq:online_fs}\end{tabular} &  &  &  &  &  \\ \hline
RetinaNet & \checkmark &  &  & 35.7 & 54.7 & 38.5 & 19.5 & 39.9 & 47.5 \\ \hline
\multirow{4}{*}{Ours} &  & \checkmark &  & 34.7 & 54.0 & 36.4 & 19.0 & 39.0 & 45.8 \\ 
 &  &  & \checkmark & 35.9 & 55.0 & 37.9 & 19.8 & 39.6 & 48.2 \\ 
 & \checkmark & \checkmark &  & 36.1 & 55.6 & 38.7 & 19.8 & 39.7 & 48.9 \\ 
 & \checkmark &  & \checkmark & \textbf{37.2} & \textbf{57.2} & \textbf{39.4} & \textbf{21.0} & \textbf{41.2} & \textbf{49.7} \\ \hline
\end{tabular}
\caption{Ablative experiments for the FSAF module on the COCO \texttt{minival}. ResNet-50 is the backbone network for all experiments in this table. We study the effect of anchor-free branches, heuristic feature selection, and online feature selection.}
\label{table:ablation}
\end{table*}

\subsection{Joint Inference and Training}
\label{section:method:training}

When plugged into RetinaNet~\cite{retinanet}, our FSAF module works jointly with the anchor-based branches, see Figure~\ref{fig:af_retina}. We keep the anchor-based branches as original, with all hyperparameters unchanged in both training and inference.

\noindent \textbf{Inference:} The FSAF module just adds a few convolution layers to the fully-convolutional RetinaNet, so the inference is still as simple as forwarding an image through the network. For anchor-free branches, we only decode box predictions from at most 1k top-scoring locations in each pyramid level, after thresholding the confidence scores by 0.05. These top predictions from all levels are merged with the box predictions from anchor-based branches, followed by non-maximum suppression with a threshold of 0.5, yielding the final detections.

\noindent \textbf{Initialization:} The backbone networks are pre-trained on ImageNet1k~\cite{imagenet}. We initialize the layers in RetinaNet as in \cite{retinanet}. For conv layers in our FSAF module, we initialize the classification layers with bias $-\log((1-\pi)/\pi)$ and a Gaussian weight filled with $\sigma = 0.01$, where $\pi$ specifies that at the beginning of training every pixel location outputs objectness scores around $\pi$. We set $\pi = 0.01$ following \cite{retinanet}. All the box regression layers are initialized with bias $b$, and a Gaussian weight filled with $\sigma = 0.01$.
We use $b = 0.1$ in all experiments. The initialization helps stabilize the network learning in the early iterations by preventing large losses.

\noindent \textbf{Optimization:} The loss for the whole network is combined losses from the anchor-free and anchor-based branches. Let $L^{ab}$ be the total loss of the original anchor-based RetinaNet. And let $L_{cls}^{af}$ and $L_{reg}^{af}$ be the total classification and regression losses of anchor-free branches, respectively. Then total optimization loss is
$
    L = L^{ab} + \lambda (L_{cls}^{af} + L_{reg}^{af})
$,
where $\lambda$ controls the weight of the anchor-free branches. We set $\lambda = 0.5$ in all experiments, although results are robust to the exact value. The entire network is trained with stochastic gradient descent (SGD) on 8 GPUs with 2 images per GPU. Unless otherwise noted, all models are trained for 90k iterations with an initial learning rate of 0.01, which is divided by 10 at 60k and again at 80k iterations. Horizontal image flipping is the only applied data augmentation unless otherwise specified. Weight decay is 0.0001 and momentum is 0.9.

\section{Experiments}
We conduct experiments on the detection track of the COCO dataset~\cite{coco}. The training data is the COCO \texttt{trainval35k} split, including all 80k images from \texttt{train} and a random 35k subset of images from the 40k \texttt{val} split. We analyze our method by ablation studies on the \texttt{minival} split containing the remaining 5k images from \texttt{val}. When comparing to the state-of-the-art methods, we report COCO AP on the \texttt{test-dev} split, which has no public labels and requires the use of the evaluation server.

\subsection{Ablation Studies}
\label{section:exp:ablation}

\begin{table}
\centering
\begin{tabular}{c|c|c c c}
\hline\hline
Backbone & Method & AP & AP$_{50}$ & \begin{tabular}[c]{@{}c@{}}Runtime\\ (ms/im)\end{tabular} \\ \hline
\multirow{3}{*}{R-50} & RetinaNet & 35.7 & 54.7 & 131 \\ 
 & \begin{tabular}[c]{@{}c@{}}Ours(FSAF)\end{tabular} & 35.9 & 55.0 & 107 \\ 
 & Ours(AB+FSAF) & 37.2 & 57.2 & 138 \\ \hline
\multirow{3}{*}{R-101} & RetinaNet & 37.7 & 57.2 & 172 \\ 
 & \begin{tabular}[c]{@{}c@{}}Ours(FSAF)\end{tabular} & 37.9 & 58.0 & 148 \\ 
 & Ours(AB+FSAF) & 39.3 & 59.2 & 180 \\ \hline
\multirow{3}{*}{X-101} & RetinaNet & 39.8 & 59.5 & 356 \\  
 & \begin{tabular}[c]{@{}c@{}}Ours(FSAF)\end{tabular} & 41.0 & 61.5 & 288 \\
 & Ours(AB+FSAF) & 41.6 & 62.4 & 362 \\ \hline
\end{tabular}
\caption{Detection accuracy and inference latency with different backbone networks on the COCO \texttt{minival}. \textbf{AB}: Anchor-based branches. \textbf{R}: ResNet. \textbf{X}: ResNeXt.}
\label{table:retina}
\end{table}

For all ablation studies, we use an image scale of 800 pixels for both training and testing. We evaluate the contribution of several important elements to our detector, including anchor-free branches, online feature selection, and backbone networks. Results are reported in Table~\ref{table:ablation} and \ref{table:retina}.

\begin{figure*}
    \centering
    \begin{subfigure}{1.03\textwidth}
        \centering
        \rotatebox[origin=l]{90}{RetinaNet}
        \includegraphics[height=3cm]{figs/5_retina.pdf}
        \includegraphics[height=3cm]{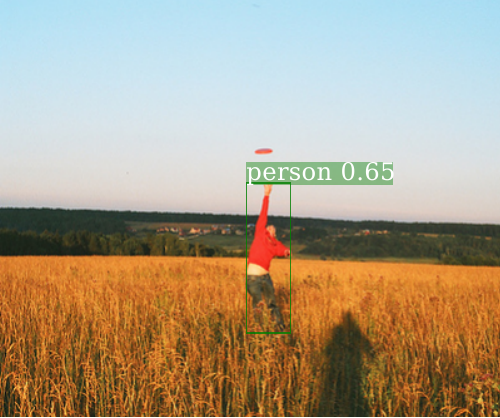}
        \includegraphics[height=3cm]{figs/2562_retina.pdf}
        \includegraphics[height=3cm]{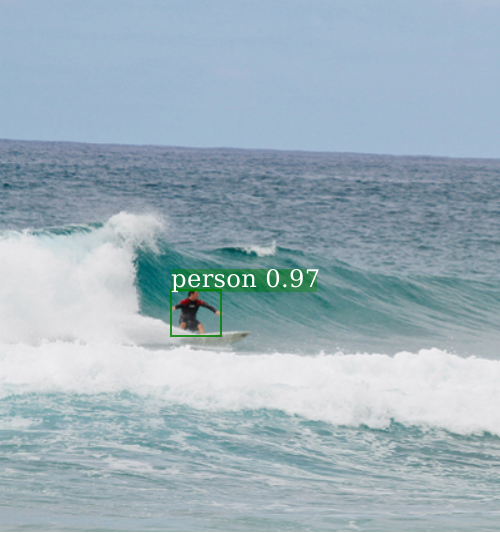}
        \includegraphics[height=3cm]{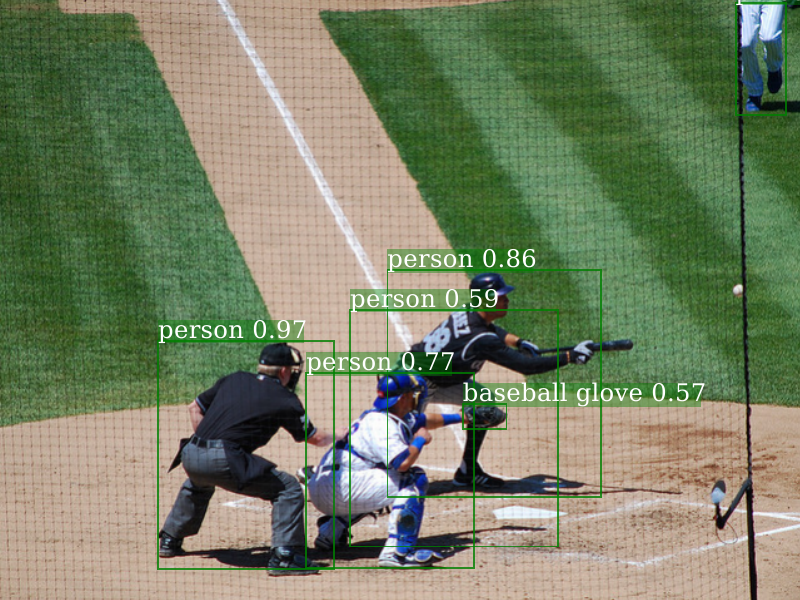}
    \end{subfigure}
    \begin{subfigure}{1.03\textwidth}
        \centering
        \rotatebox[origin=l]{90}{Ours}
        \includegraphics[height=3cm]{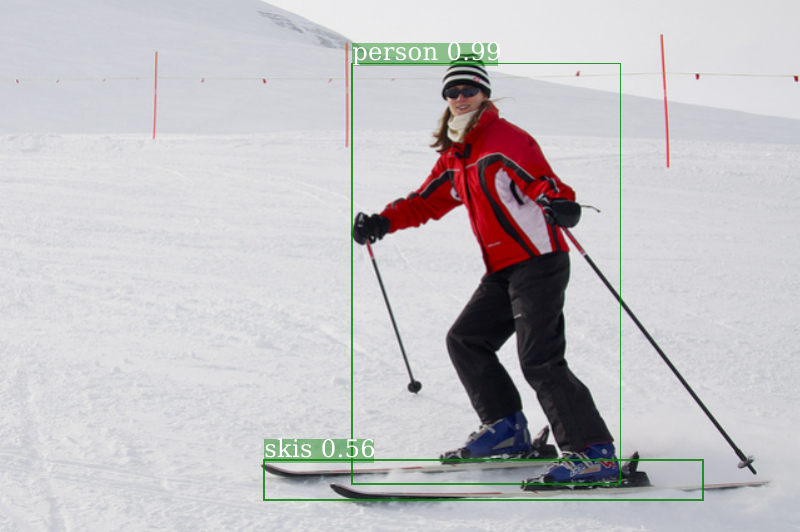}
        \includegraphics[height=3cm]{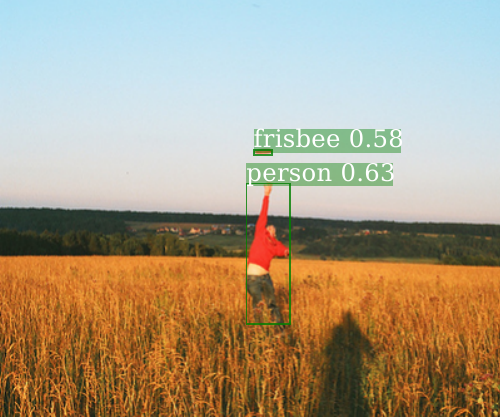}
        \includegraphics[height=3cm]{figs/2562_ours.pdf}
        \includegraphics[height=3cm]{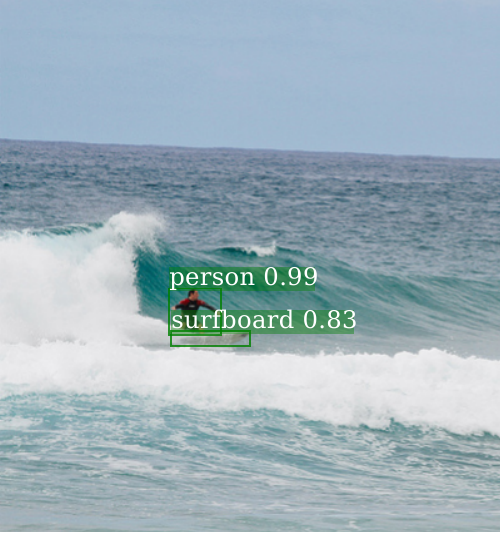}
        \includegraphics[height=3cm]{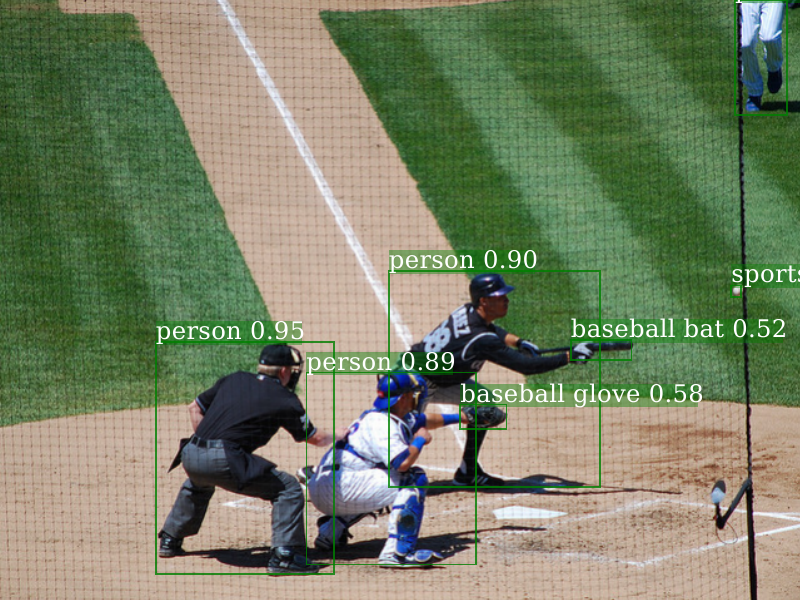}
    \end{subfigure}
    \caption{More qualitative comparison examples between anchor-based RetinaNet (top, Table~\ref{table:ablation} 1st entry) and our detector with additional FSAF module (bottom, Table~\ref{table:ablation} 5th entry). Both are using ResNet-50 as backbone. Our FSAF module helps finding more challenging objects.}
    \label{fig:ablation:compare}
\end{figure*}

\textbf{Anchor-free branches are necessary.} We first train two detectors with \textit{only} anchor-free branches, using two feature selection methods respectively (Table~\ref{table:ablation} 2nd and 3rd entries). It turns out anchor-free branches only can already achieve decent results. When jointly optimized with anchor-based branches, anchor-free branches help learning instances which are hard to be modeled by anchor-based branches, leading to improved AP scores (Table~\ref{table:ablation} 5th entry). Especially the AP$_{50}$, AP$_S$ and AP$_L$ scores increase by 2.5\%, 1.5\%, and 2.2\% respectively with online feature selection. To find out what kinds of objects the FSAF module can detect, we show some qualitative results of the head-to-head comparison between RetinaNet and ours in Figure~\ref{fig:ablation:compare}. Clearly, our FSAF module is better at finding challenging instances, such as tiny and very thin objects which are not well covered by anchor boxes. 

\begin{figure*}
    \centering
    \includegraphics[height=3cm]{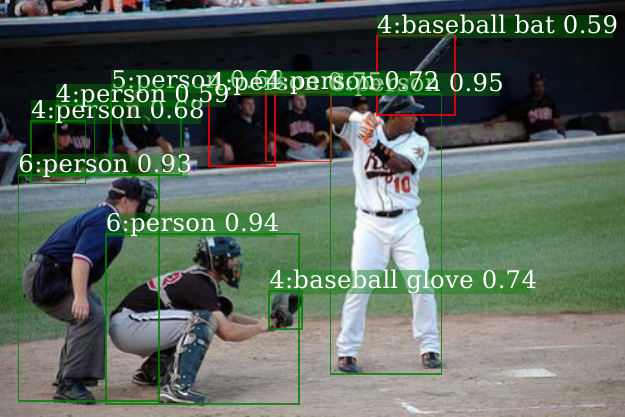}
    \includegraphics[height=3cm]{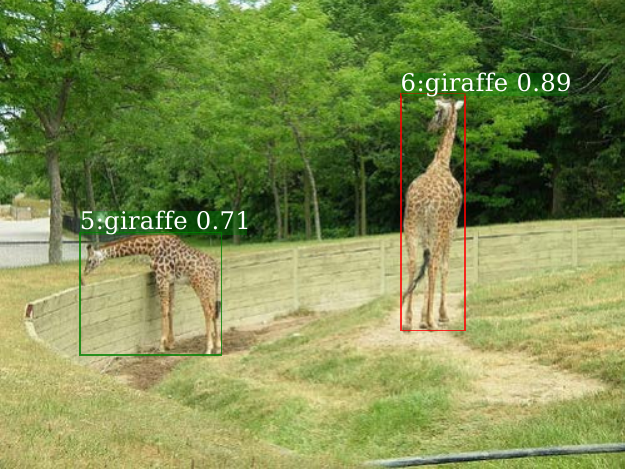}
    \includegraphics[height=3cm]{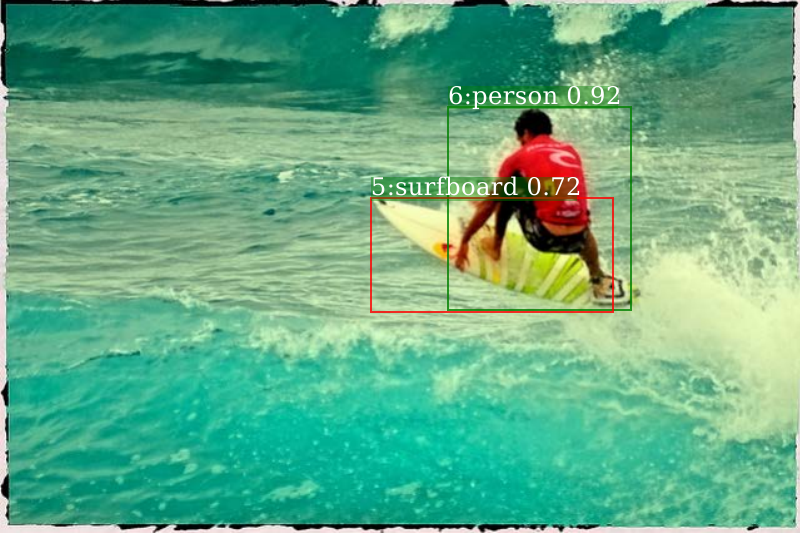}
    \includegraphics[height=3cm]{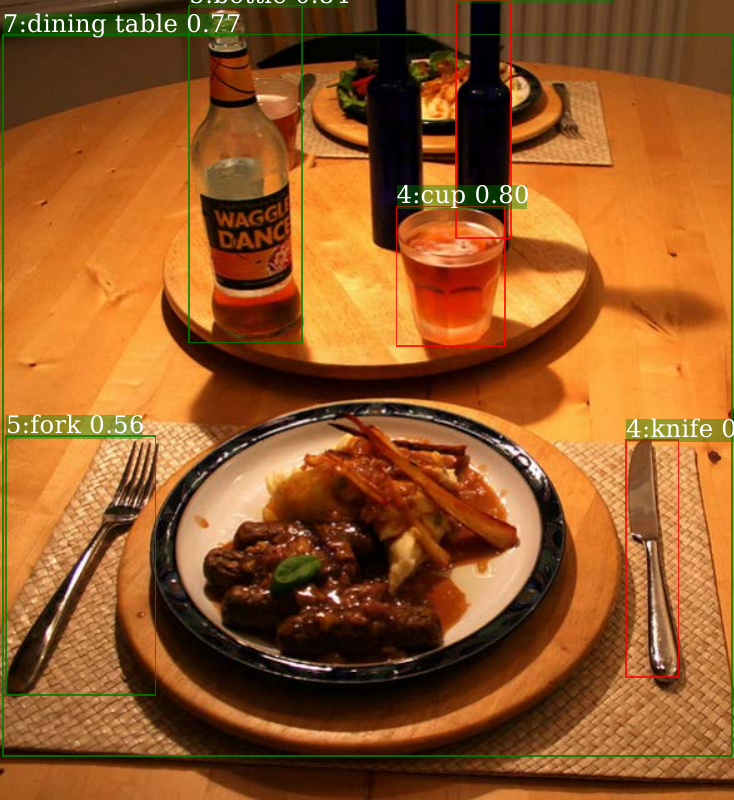}
    \includegraphics[height=3cm]{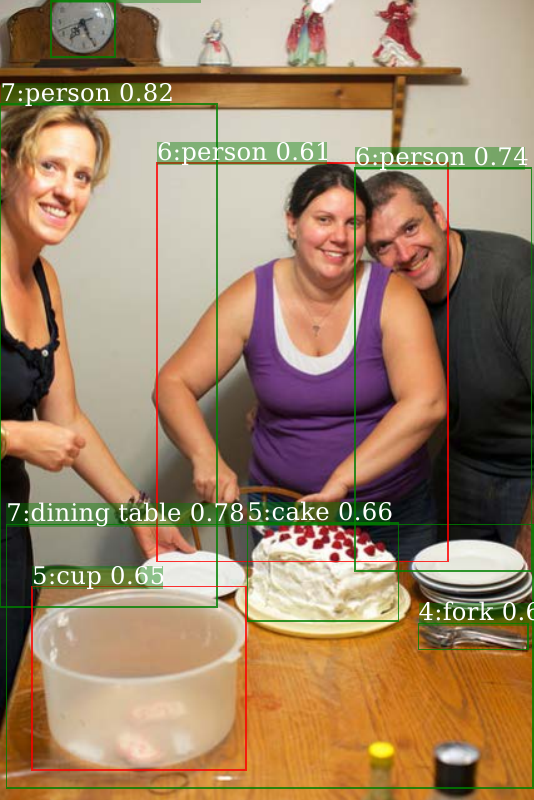}
    \includegraphics[height=3cm]{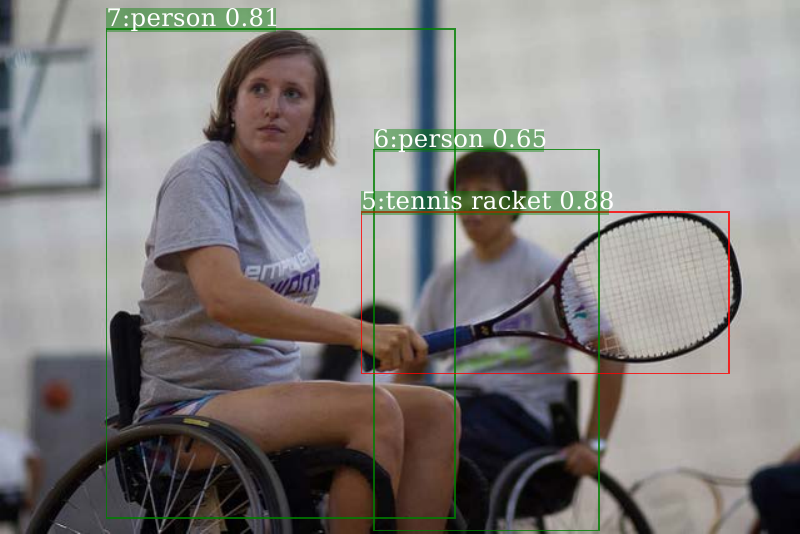}
    \includegraphics[height=3cm]{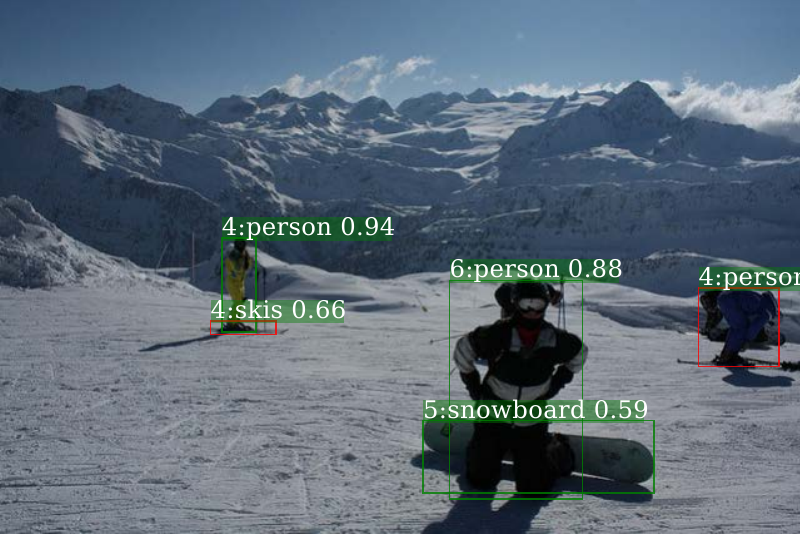}
    \includegraphics[height=3cm]{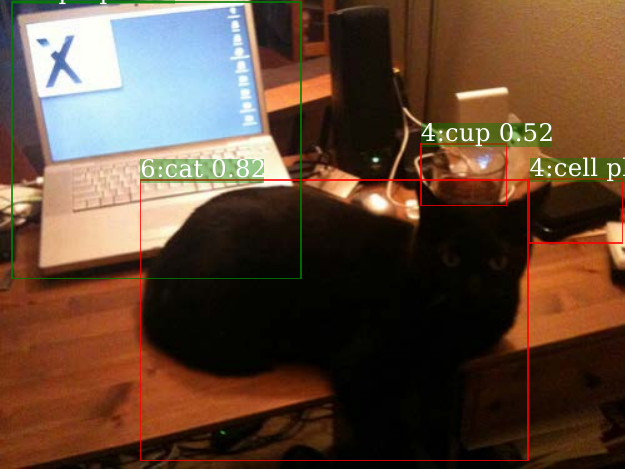}
    \includegraphics[height=3cm]{figs/4694.pdf}
    \includegraphics[height=3cm]{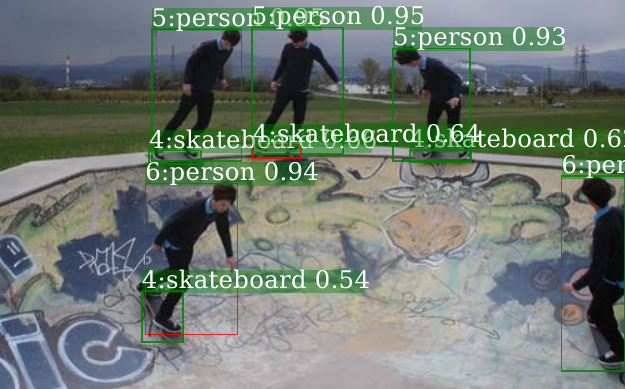}
    \includegraphics[height=3cm]{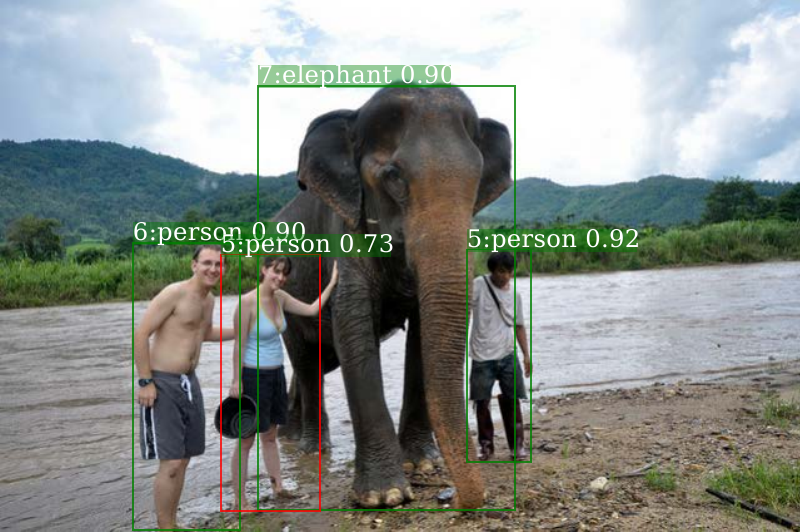}
    \includegraphics[height=3cm]{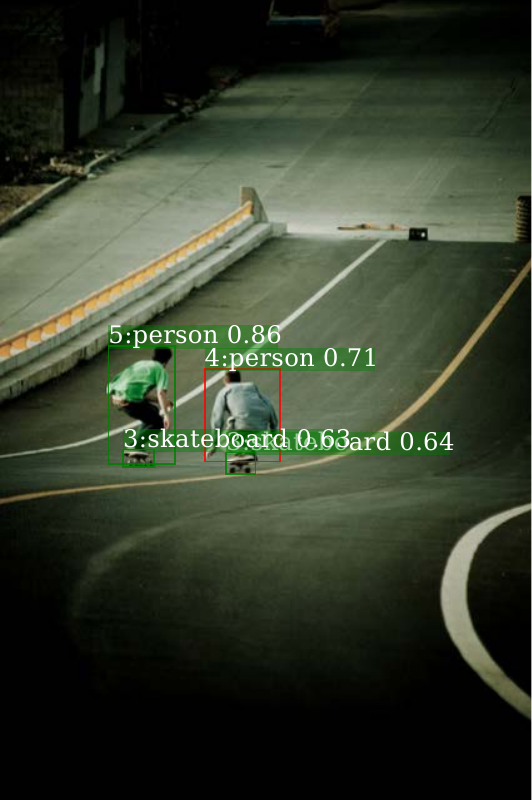}
    \includegraphics[height=3cm]{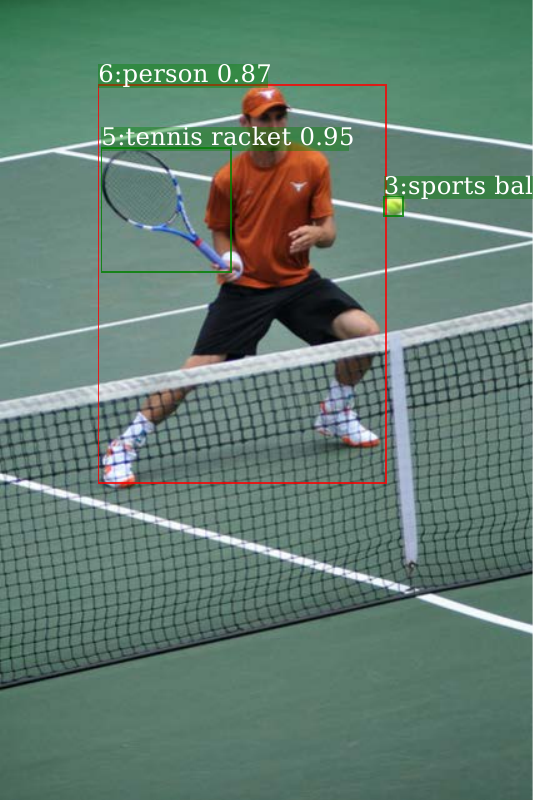}
    \caption{Visualization of online feature selection from anchor-free branches. The number before the class name is the pyramid level that detects the instance. We compare this level with the level to which as if this instance is assigned in the anchor-based branches, and use \textit{red} to indicate the disagreement and \textit{green} for agreement.}
    \label{fig:vis_feat}
\end{figure*}

\begin{table*}[t]
\centering
\begin{tabular}{c|c|c c c c c c}
\hline \hline
Method & Backbone & AP & AP$_{50}$ & AP$_{75}$ & AP$_{S}$ & AP$_{M}$ & AP$_{L}$\\
\hline \hline
\multicolumn{8}{l}{\textbf{Multi-shot detectors}} \\ \hline
CoupleNet~\cite{couplenet} & \multirow{6}{*}{ResNet-101} & 34.4 & 54.8 & 37.2 & 13.4 & 38.1 & 50.8 \\ 
Faster R-CNN+++~\cite{fasterrcnn} & & 34.9 & 55.7 & 37.4 & 15.6 & 38.7 & 50.9 \\
Faster R-CNN w/ FPN~\cite{fpn} & & 36.2 & 59.1 & 39.0 & 18.2 & 39.0 & 48.2 \\
Regionlets~\cite{regionlets} & & 39.3 & 59.8 & n/a & 21.7 & 43.7 & 50.9 \\
Fitness NMS~\cite{fitnessnms} & & 41.8 & 60.9 & 44.9 & 21.5 & 45.0 & 57.5 \\
Cascade R-CNN~\cite{cascade} & & 42.8 & 62.1 & 46.3 & 23.7 & 45.5 & 55.2 \\ \hline
Deformable R-FCN~\cite{dcn} & \multirow{2}{*}{Aligned-Inception-ResNet} & 37.5 & 58.0 & n/a & 19.4 & 40.1 & 52.5 \\
Soft-NMS~\cite{softnms} & & 40.9 & 62.8 & n/a & 23.3 & 43.6 & 53.3 \\ \hline
Deformable R-FCN + SNIP~\cite{snip} & DPN-98 & 45.7 & 67.3 & 51.1 & 29.3 & 48.8 & 57.1 \\
\hline \hline
\multicolumn{8}{l}{\textbf{Single-shot detectors}} \\ \hline
YOLOv2~\cite{yolo9000} & DarkNet-19 & 21.6 & 44.0 & 19.2 & 5.0 & 22.4 & 35.5 \\ \hline
SSD513~\cite{ssd} & \multirow{8}{*}{ResNet-101} & 31.2 & 50.4 & 33.3 & 10.2 & 34.5 & 49.8 \\ 
DSSD513~\cite{dssd} &  & 33.2 & 53.3 & 35.2 & 13.0 & 35.4 & 51.1 \\ 
RefineDet512~\cite{refinedet} (single-scale) &  & 36.4 & 57.5 & 39.5 & 16.6 & 39.9 & 51.4 \\ 
RefineDet~\cite{refinedet} (multi-scale) &  & 41.8 & 62.9 & 45.7 & 25.6 & 45.1 & \textbf{54.1} \\ 
RetinaNet800~\cite{retinanet} &  & 39.1 & 59.1 & 42.3 & 21.8 & 42.7 & 50.2 \\ 
GHM800~\cite{ghm} &  & 39.9 & 60.8 & 42.5 & 20.3 & 43.6 & \textbf{54.1} \\
\textbf{Ours800} (single-scale) &  & 40.9 & 61.5 & 44.0 & 24.0 & 44.2 & 51.3 \\ 
\textbf{Ours} (multi-scale) &  & \textbf{42.8} & \textbf{63.1} & \textbf{46.5} & \textbf{27.8} & \textbf{45.5} & 53.2 \\ \hline
CornerNet511~\cite{cornernet} (single-scale) & \multirow{2}{*}{Hourglass-104} & 40.5 & 56.5 & 43.1 & 19.4 & 42.7 & 53.9 \\ 
CornerNet~\cite{cornernet} (multi-scale) &  & 42.1 & 57.8 & 45.3 & 20.8 & 44.8 & 56.7 \\ \hline
GHM800~\cite{ghm} & \multirow{3}{*}{ResNeXt-101} & 41.6 & 62.8 & 44.2 & 22.3 & 45.1 & \textbf{55.3} \\
\textbf{Ours800} (single-scale) &  & 42.9 & 63.8 & 46.3 & 26.6 & 46.2 & 52.7 \\ 
\textbf{Ours} (multi-scale) &  & \textbf{44.6} & \textbf{65.2} & \textbf{48.6} & \textbf{29.7} & \textbf{47.1} & 54.6 \\ \hline
\end{tabular}
\caption{Object detection results of our best  \textit{single} model with the FSAF module vs. state-of-the-art single-shot and multi-shot detectors on the COCO \texttt{test-dev}. }
\label{table:sota}
\end{table*}

\textbf{Online feature selection is essential.} 
As stated in Section~\ref{section:method:feature_selection}, we can select features in anchor-free branches either based on heuristics just like the anchor-based branches, or based on instance content. It turns out selecting the right feature to learn plays a fundamental role in detection. Experiments show that anchor-free branches with heuristic feature selection (Eqn.~\eqref{eq:naive_fs}) only are not able to compete with anchor-based counterparts due to less learnable parameters. But with our online feature selection (Eqn.~\eqref{eq:online_fs}), the AP is improved by \textbf{1.2\%} (Table~\ref{table:ablation} 3rd vs 2nd entries), which overcomes the parameter disadvantage.
Additionally, Table~\ref{table:ablation} 4th and 5th entries further confirm that our online feature selection is essential for anchor-free and anchor-based branches to work well together.


\textbf{How is optimal feature selected?} In order to understand the optimal pyramid level selected for instances, we visualize some qualitative detection results from only the anchor-free branches in Figure~\ref{fig:vis_feat}. 
The number before the class name indicates the feature level that detects the object. It turns out the online feature selection actually follows the rule that upper levels select larger instances, and lower levels are responsible for smaller instances, which is the same principle in anchor-based branches. However, there are quite a few exceptions, \ie online feature selection chooses pyramid levels different from the choices of anchor-based branches. We label these exceptions as red boxes in Figure~\ref{fig:vis_feat}. Green boxes indicate agreement between the FSAF module and anchor-based branches. By capturing these exceptions, our FSAF module can use better features to detect challenging objects. 



\textbf{FSAF module is robust and efficient.} We also evaluate the effect of backbone networks to our FSAF module in terms of accuracy and speed. Three backbone networks include ResNet-50, ResNet-101~\cite{resnet}, and ResNeXt-101~\cite{resnext}. Detectors run on a single Titan X GPU with CUDA 9 and CUDNN 7 using a batch size of 1. Results are reported in Table~\ref{table:retina}. We find that our FSAF module is robust to various backbone networks. The FSAF module by itself is already better and faster than anchor-based RetinaNet. On ResNeXt-101, the FSAF module outperforms anchor-based counterparts by \textbf{1.2\%} AP while being \textbf{68ms} faster. When applied jointly with anchor-based branches, our FSAF module consistently offers considerable improvements. This also suggests that \textit{anchor-based branches are not utilizing the full power of backbone networks.} Meanwhile, our FSAF module introduces marginal computation cost to the whole network, leading to negligible loss of inference speed. Especially, we improve RetinaNet by \textbf{1.8\%} AP on ResNeXt-101 with only \textbf{6ms} additional inference latency. 

\subsection{Comparison to State of the Art}

We evaluate our final detector on the COCO \texttt{test-dev} split to compare with recent state-of-the-art methods. Our final model is RetinaNet with the FSAF module, i.e. anchor-based branches plus the FSAF module. 
The model is trained using scale jitter over scales \{640, 672, 704, 736, 768, 800\} and for $1.5\times$ longer than the models in Section~\ref{section:exp:ablation}. The evaluation includes single-scale and multi-scale versions, where single-scale testing uses an image scale of 800 pixels and multi-scale testing applies test time augmentations. Test time augmentations are testing over scales \{400, 500, 600, 700, 900, 1000, 1100, 1200\} and horizontal flipping on each scale, following Detectron~\cite{detectron}. All of our results are from single models \textit{without} ensemble. 

Table~\ref{table:sota} presents the comparison. With ResNet-101, our detector is able to achieve competitive performance in both single-scale and multi-scale scenarios. Plugging in ResNeXt-101-64x4d further improves AP to \textbf{44.6\%} , which outperforms previous state-of-the-art single-shot detectors by a large margin.

\section{Conclusion}
This work identifies heuristic feature selection as the primary limitation for anchor-based single-shot detectors with feature pyramids. To address this, we propose FSAF module which applies online feature selection to train anchor-free branches in the feature pyramid. It significantly improves strong baselines with tiny inference overhead and outperforms recent state-of-the-art single-shot detectors.


{\small
\bibliographystyle{ieee}
\bibliography{egbib}
}
\end{document}